\documentclass[runningheads]{llncs}
\usepackage{graphicx}
\usepackage{booktabs}
\usepackage{array}
\usepackage{multirow}
\usepackage{float}
\usepackage{subfigure}
\usepackage{booktabs}
\begin{document}
\bibliographystyle{unsrt}
\title{Lidar-based Object Classification with Explicit Occlusion Modeling}

\author{Xiaoxiang Zhang\inst{1} \and Hao Fu\inst{1} \and
Bin Dai\inst{1,2,3}}

\institute{College of Intelligence Science and Technology, National University of Defense Technology, Changsha 410073, China; zhangxiaoxiang17@qq.com; fuhao@nudt.edu.cn \and
Unmanned Systems Research Center, National Innovation Institute of Defense Technology,Beijing 100071, China
\and Correspondence:bindai.cs@gmail.com}

\maketitle              

\begin{abstract}
LIDAR is one of the most important sensors for Unmanned Ground Vehicles (UGV). Object detection and classification based on lidar point cloud is a key technology for UGV. In object detection and classification, the mutual occlusion between neighboring objects is an important factor affecting the accuracy. In this paper, we consider occlusion as an intrinsic property of the point cloud data. We propose a novel approach that explicitly model the occlusion. The occlusion property is then taken into account in the subsequent classification step. We perform experiments on the KITTI dataset. Experimental results indicate that by utilizing the occlusion property that we modeled, the classifier obtains much better performance.

\keywords{Object classification \and LIDAR \and UGV \and Occlusion.}
\end{abstract}

\section{Introduction}
LIDAR is one of the most popular sensors for unmanned vehicle due to its highly precise range measurements. Object detection and classification based on lidar point cloud is an extremely important technology for unmanned vehicle. However, the sparseness of the lidar point cloud and the mutual occlusion between neighboring objects poses significant challenges for object detection and classification algorithm. Fig.1 is a typical traffic scene. We could observe a lot of occlusions occurred in this figure.
\begin{figure}
\begin{center}
\includegraphics[width=1.0\linewidth]{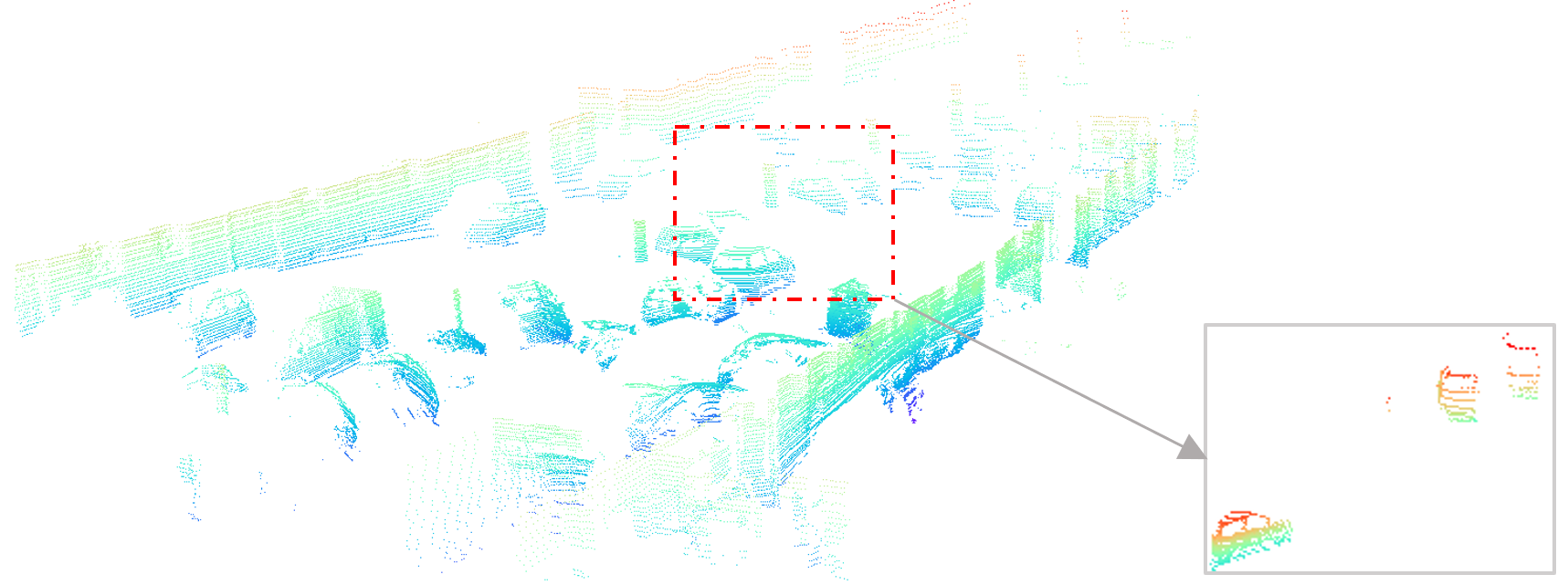}
\end{center}
   \caption{In a typical traffic scenario, it is common to see the mutual occlusion between neighboring objects. The lidar point cloud of the object to be classified is often incomplete and fragmented, which could easily result in wrong classification results.}
\label{fig:long}
\label{fig:onecol}
\end{figure}

Ideally, the lidar point cloud corresponding to an object should be relatively complete and fully reflect the spatial distribution characteristics of objects. However, due to the mutual occlusion of neighboring objects, the object point cloud is usually incomplete which may result in the wrong classification of the object. 

An illustrative example is shown in Fig.2. In the training phase, many positive samples, including sample A and B as shown in the top row of Fig.2, are fed into the classifier. It is seen that sample B is occluded by another obstacle, making its point cloud incomplete. The classifier is then trained to adapt to this intra-class variation. In the testing phase, the classifier encounters two samples, C and D. Among them, sample D is a true positive while C is composed of two small objects, E and F. The classifier will encounter difficulties in distinguishing C from D, and it is very likely to classify C as a false positive or classify D as a false negative.

\begin{figure*}[t]
\includegraphics[width=\textwidth]{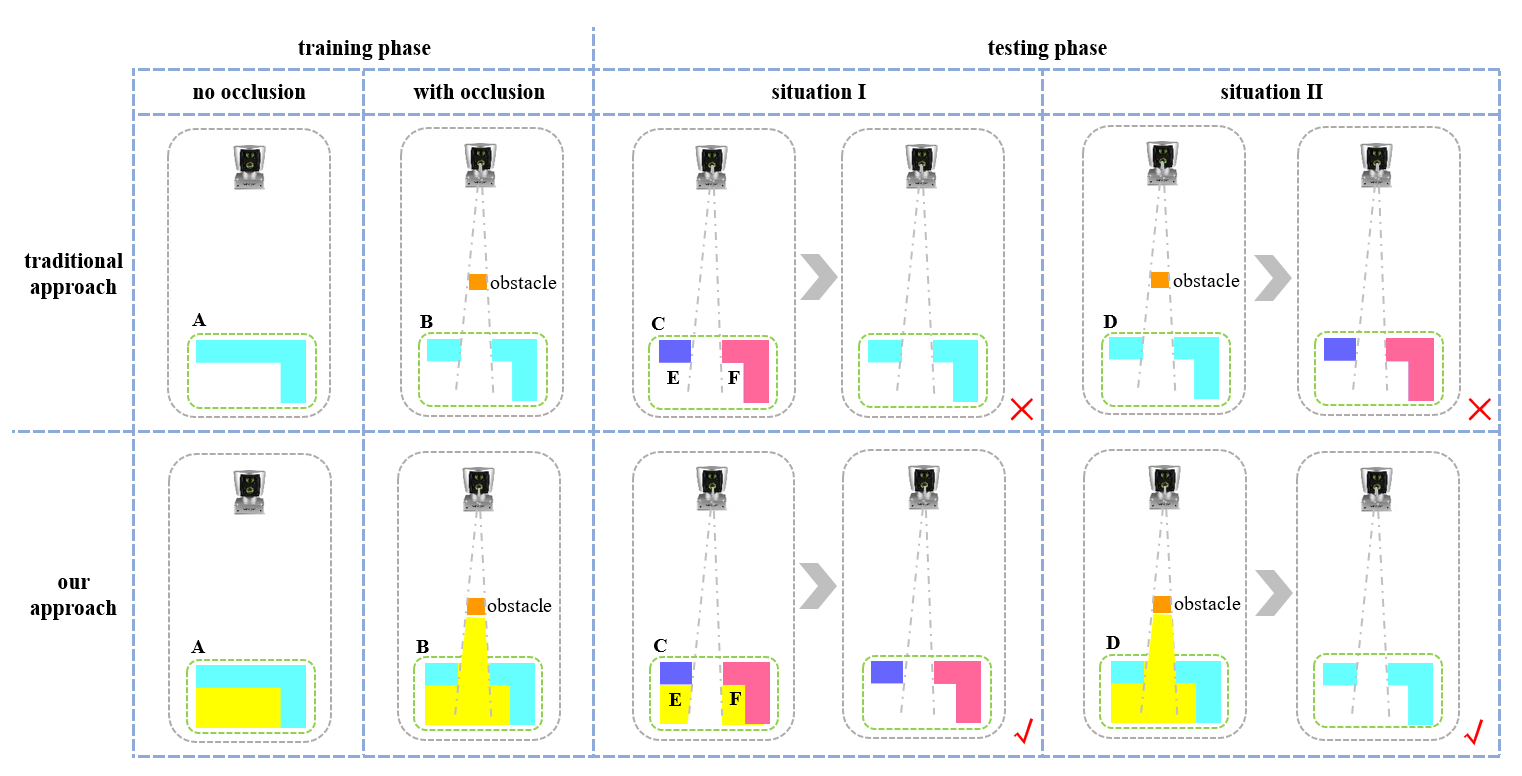}
   \caption{In the training phase of the traditional approach, many positive samples, including sample A and B as shown in the top row, are fed into the classifier. Sample B is occluded by another obstacle, making its point cloud incomplete. In the testing phase, the classifier encounters two samples, C and D. Sample C is likely to be classified as false positive while sample D is likely to be classified as a false negative. In our approach, we add a pre-processing step that computes the occlusion property of the point cloud before the classification module. As shown in the bottom row, the occluded area is colored in yellow. With the help of the occlusion area, the classifier can now easily distinguish object C from D, thus both the false positive rate and false negative rate might be reduced.}
\label{fig:long}
\label{fig:onecol}
\end{figure*}

In this paper, we consider occlusion as an intrinsic property of the point cloud data. The occlusion area could be accurately computed by considering the relative position between the LIDAR itself and each detected LIDAR point using ray-casting technique\cite{roth1982ray}. Therefore, we add a pre-processing step to add the occlusion property to the point cloud before any further processing. As shown in the bottom row of Fig.2, the occluded area is colored in yellow. With the help of the occlusion area, the classifier can now easily distinguish object C from D. Therefore, both the false positive rate and false negative rate might be reduced. 

We test our approach on the KITTI dataset. We choose to use PointNet\cite{qi2017pointnet} as the basic classifier. We modified PointNet to enable it to utilize the occlusion property. Experimental results show that our method obtains a significant improvement compared to the original PointNet, both in the overall classification accuracy and per-class classification accuracy.

\section{Related Work}
There has been a large literature on object detection approaches based on point cloud. Petrovskaya et al. proposed an object detection algorithm based on object geometry and motion model~\cite{petrovskaya2009model1,petrovskaya2009model,petrovskaya2009efficient} , and used Bayesian filters to estimate the model parameters. Himmelsbach et al. extracted the geometric features of point cloud using the point feature histogram~\cite{himmelsbach2009real,wang2003online} , and then used SVM to classify the object. Built on the work of~\cite{petrovskaya2009model1,petrovskaya2009model,petrovskaya2009efficient} , Wojke et al. ~\cite{wojke2012moving} proposed an object detection algorithm based on the combination of line features and angular features. Cheng et al. proposed to use histogram features for object detection and recognition~\cite{cheng2014robust} .

Recently, deep learning based approaches have become popular due to its outstanding performance. MV3D~\cite{chen2017multi} firstly projects point cloud onto the bird's eye view and then trains a region proposal network (RPN) for generating 3D bounding box proposals. However, MV3D does not perform well in detecting small objects such as pedestrians and cyclists. VoxelNet~\cite{zhou2018voxelnet} is an end-to-end object detection framework. It divides the point cloud into equally spaced three-dimensional voxels and then transforms the points in each voxel into a uniform feature representation through the newly introduced Voxel Feature Coding (VFE) layer. The point cloud is then encoded as a volume representation to perform the detection and classification. Different from those previous approaches that rely on a mid-level representation, such as the image grids or the 3D voxels, Qi et al. proposed a new type of network called PointNet~\cite{qi2017pointnet} that works directly on the original point cloud. PointNet is a unified framework that can be applied to object classification, part segmentation and scene semantic parsing. It obtains competitive results on several 3D object classification benchmarks. 

For occlusion handling, there have been several works~\cite{Wang_2018_CVPR,Zhang_2018_ECCV,Baque_2017_ICCV,Edward_2014} trying to directly predict the occlusion mask. However, most of these works are image-based approaches. There has been little work on lidar-based approaches that directly models occlusion and utilize the occlusion property to aid the classification tasks.

\section{The Proposed Approach}
\subsection{Point Cloud Definition}
A point cloud is represented as a set of three dimensional points $\left\{P_i | i = 1, ..., n\right\}$, where each point $P_i$ is a vector of $(x, y, z)$.

We define the object point cloud data within the object bounding box as the object point cloud $P_{raw}=\left\{P_{raw}^{1}, P_{raw}^{2},...,P_{raw}^{n}\right\}$. The point cloud outside the object bounding box is defined as the obstacle point cloud $ P_{ob}=\left\{P_{ob}^{1}, P_{ob}^{2},...,P_{ob}^{m}\right\} $. The obstacle point cloud will block the lidar ray from passing through it, thus resulting in an incomplete object point cloud. In Fig.3, we can see that the object point cloud is divided into two parts. The occlusion area generated by the point cloud using the ray-casting technique is defined as the occluded point cloud $ P_{oc}=\left\{P_{oc}^{1}, P_{oc}^{2},...,P_{oc}^{k}\right\} $, and is colored in yellow and pink respectively. 

\begin{figure}
\begin{center}
\includegraphics[width=0.8\linewidth]{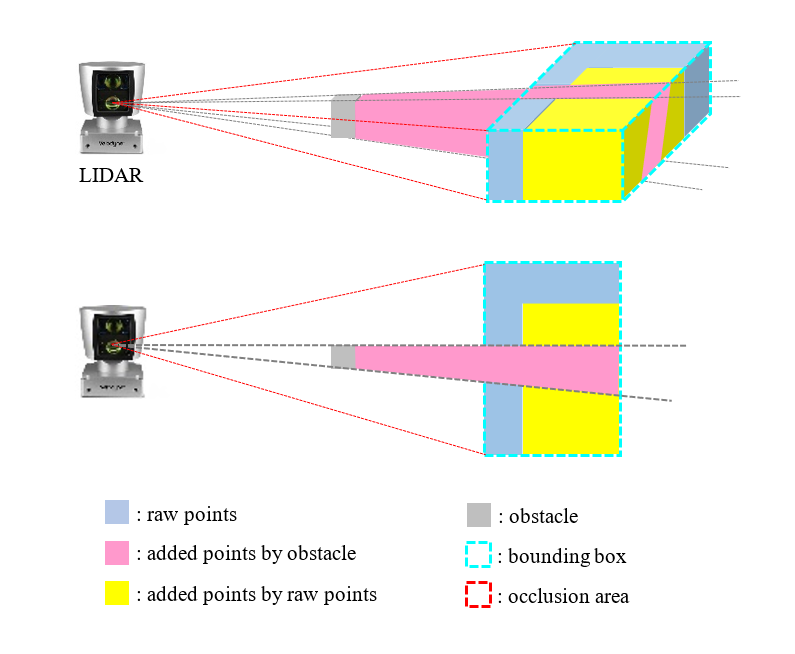}
\end{center}
   \caption{Point cloud definition. The top figure is the 3D-view and the bottom figure is the corresponding birds-eye view. The gray cube represents the obstacle point cloud. The object point cloud is colored in blue. The occlusion area generated by the point cloud is colored in pink and yellow.}
\label{fig:long}
\label{fig:onecol}
\end{figure}

\subsection{Occlusion Area Modeling}
For each point $P_{ob}^{i}=\left(x_{ob}^{i},y_{ob}^{i},z_{ob}^{i}\right) , i=1,2,3,...,m$ of the obstacle point cloud and each point $P_{raw}^{j}=\left(x_{raw}^{j},y_{raw}^{j},z_{raw}^{j}\right) , j=1,2,3,...,n$ of the raw object point cloud, we use the ray-casting technique to model the occlusion. We define the position of the LIDAR as the origin $O$. For each point $P_{ob}^{i}$ and $P_{raw}^{j}$, we add occluded points $ P_{oc}^{l}$ along the direction of $O$ to $P_{ob}^{i}$ or $P_{raw}^{j}$ at a fixed step. The occluded points are added until their height is below the ground plane. The ground plane is estimated by using a block recursive Gaussian process regression algorithm~\cite{CTT}.

For each point $P_{ob}^{i}$ of the obstacle point cloud:
\begin{equation} 
\frac{L(OP_{oc}^{l_1})}{L(OP_{ob}^{i})}=\frac{x_{oc}^{l_1}}{x_{ob}^{i}}=\frac{y_{oc}^{l_1}}{y_{ob}^{i}}=\frac{z_{oc}^{l_1}}{z_{ob}^{i}}
\end{equation} 
\begin{equation}
L(OP_{oc}^{l_1})=L(OP_{ob}^{i})+k_1s
\end{equation}

For each point $P_{raw}^{j}$ of the object point cloud:
\begin{equation} 
\frac{L(OP_{oc}^{l_2})}{L(OP_{raw}^{j})}=\frac{x_{oc}^{l_2}}{x_{raw}^{j}}=\frac{y_{oc}^{l_2}}{y_{raw}^{j}}=\frac{z_{oc}^{l_2}}{z_{raw}^{j}}
\end{equation} 
\begin{equation}
L(OP_{oc}^{l_2})=L(OP_{raw}^{j})+k_2s
\end{equation}
where $k_1$, $k_2$ are positive integer, $s$ is the step size ( in our experiment we set $s=0.3m$). The function $L(x)$ represents the distance from the point $P$ to the origin $O$.

\begin{figure}[htbp]
\centering
\begin{minipage}[t]{0.13\linewidth}
\centerline{\includegraphics[width=0.8in]{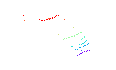}}
\end{minipage}
\quad
\begin{minipage}[t]{0.13\linewidth}
\centerline{\includegraphics[width=0.8in]{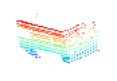}}
\end{minipage}
\quad
\begin{minipage}[t]{0.13\linewidth}
\centerline{\includegraphics[width=0.8in]{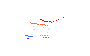}}
\end{minipage}
\quad
\begin{minipage}[t]{0.13\linewidth}
\centerline{\includegraphics[width=0.8in]{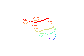}}
\end{minipage}
\quad
\begin{minipage}[t]{0.13\linewidth}
\centerline{\includegraphics[width=0.8in]{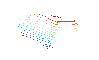}}
\end{minipage}
\quad
\begin{minipage}[t]{0.13\linewidth}
\centerline{\includegraphics[height=0.6in]{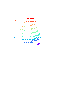}}
\end{minipage}
\quad

\begin{minipage}[t]{0.13\linewidth}
\centerline{\includegraphics[width=0.8in]{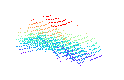}}
\centerline{(a)}
\end{minipage}
\quad
\begin{minipage}[t]{0.13\linewidth}
\centerline{\includegraphics[width=0.8in]{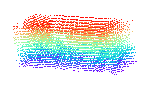}}
\centerline{(b)}
\end{minipage}
\quad
\begin{minipage}[t]{0.13\linewidth}
\centerline{\includegraphics[width=0.8in]{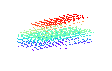}}
\centerline{(c)}
\end{minipage}
\quad
\begin{minipage}[t]{0.13\linewidth}
\centerline{\includegraphics[width=0.8in]{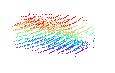}}
\centerline{(d)}
\end{minipage}
\quad
\begin{minipage}[t]{0.13\linewidth}
\centerline{\includegraphics[width=0.8in]{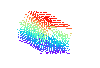}}
\centerline{(e)}
\end{minipage}
\quad
\begin{minipage}[t]{0.13\linewidth}
\centerline{\includegraphics[height=0.6in]{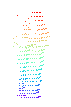}}
\centerline{(f)}
\end{minipage}
\quad

\begin{minipage}[t]{0.13\linewidth}
\centerline{\includegraphics[width=0.8in]{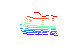}}
\end{minipage}
\quad
\begin{minipage}[t]{0.13\linewidth}
\centerline{\includegraphics[width=0.8in]{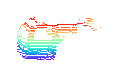}}
\end{minipage}
\quad
\begin{minipage}[t]{0.13\linewidth}
\centerline{\includegraphics[width=0.8in]{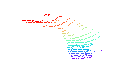}}
\end{minipage}
\quad
\begin{minipage}[t]{0.13\linewidth}
\centerline{\includegraphics[width=0.8in]{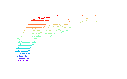}}
\end{minipage}
\quad
\begin{minipage}[t]{0.13\linewidth}
\centerline{\includegraphics[width=0.8in]{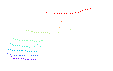}}
\end{minipage}
\quad
\begin{minipage}[t]{0.13\linewidth}
\centerline{\includegraphics[width=0.8in]{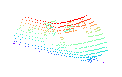}}
\end{minipage}
\quad

\begin{minipage}[t]{0.13\linewidth}
\centerline{\includegraphics[width=0.8in]{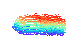}}
\centerline{(g)}
\end{minipage}
\quad
\begin{minipage}[t]{0.13\linewidth}
\centerline{\includegraphics[width=0.8in]{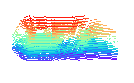}}
\centerline{(h)}
\end{minipage}
\quad
\begin{minipage}[t]{0.13\linewidth}
\centerline{\includegraphics[width=0.8in]{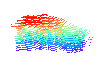}}
\centerline{(i)}
\end{minipage}
\quad
\begin{minipage}[t]{0.13\linewidth}
\centerline{\includegraphics[width=0.8in]{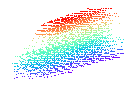}}
\centerline{(j)}
\end{minipage}
\quad
\begin{minipage}[t]{0.13\linewidth}
\centerline{\includegraphics[width=0.8in]{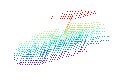}}
\centerline{(k)}
\end{minipage}
\quad
\begin{minipage}[t]{0.13\linewidth}
\centerline{\includegraphics[width=0.8in]{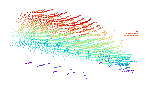}}
\centerline{(l)}
\end{minipage}
\quad
\centering
\caption{The comparison of the object point cloud with and without the occluded points. The first row and the third row are the raw object point cloud without occluded points. The second and the fourth row are the new object point cloud with occluded points. We can clearly see that the object point cloud with occluded points is more complete compared with the original one.}
\end{figure}

To distinguish the added occluded point cloud from the original point cloud, we add a new dimension named `occluded' to the original point cloud data, expanding the point cloud dimension from three dimensional $(x,y,z)$ to four dimensional $(x,y,z,o)$. We set the occlusion property of original point clouds to 0 and set the occlusion property of newly added occlusion points to 1. We use our approach to add the occluded point cloud for both the object and the obstacle point cloud. 

We show the comparison of the object point cloud with and without the occluded points in Fig.4. The first row and the third row are the raw object point cloud without occluded points. The second and the fourth row are the new object point cloud with occluded points. It is obvious that the object point cloud with occluded points is more complete than the raw object point cloud.

\subsection{Deep Learning Based Point Cloud Classification Approach}
We choose to use PointNet as the classification approach. PointNet\cite{qi2017pointnet} proposed by Qi et. al, is an method that directly processes the original point cloud. PointNet mainly consists of several transformation layers and several Multi-Layer Perceptron (MLP) blocks. The first layer of PointNet takes $n$ points as input and learns a ${D}\times{D}$ transformation matrix through the T-Net learning, where $D$ represents the feature dimension.

The transformed data then goes through several Multi-Layer Perceptron(MLP) blocks shared by each point, an intermediate max pooling layer, a spatial transformation layer and two fully connected layers. The initial value of the spatial transformation matrix is set to an identity matrix. Except for the last layer, ReLU and Batch Normalization are applied to all other layers. 

MLP of PointNet is implemented by the convolution of shared weights. The convolution kernel of the first layer is ${1}\times{3}$, and the subsequent convolution kernel size is ${1}\times{1}$. 

\subsection{Deep Learning Based Point Cloud Classification Approach With Occlusion Modeling}
Based on the original PointNet network, we make some modifications to utilize the occlusion property proposed in this paper. We expand the input data from 3D to 4D, i.e. ${n}\times{4}$ in order to enable PointNet to process new formats of point cloud data. For the ${D}\times{D}$ transformation matrix obtained by the T-Net learning, we have also modified them so that the feature dimension of the new transformation matrix becomes ${4}\times{4}$. 

In the subsequent module, we have also made appropriate modifications to the network. The size of the convolution kernel of the MLP is modified to ${1}\times{4}$ according to the input data dimension, and the output dimension of the last layer is set to the number of classes.
\begin{figure*}[h]
\begin{center}
\includegraphics[width=1.0\linewidth]{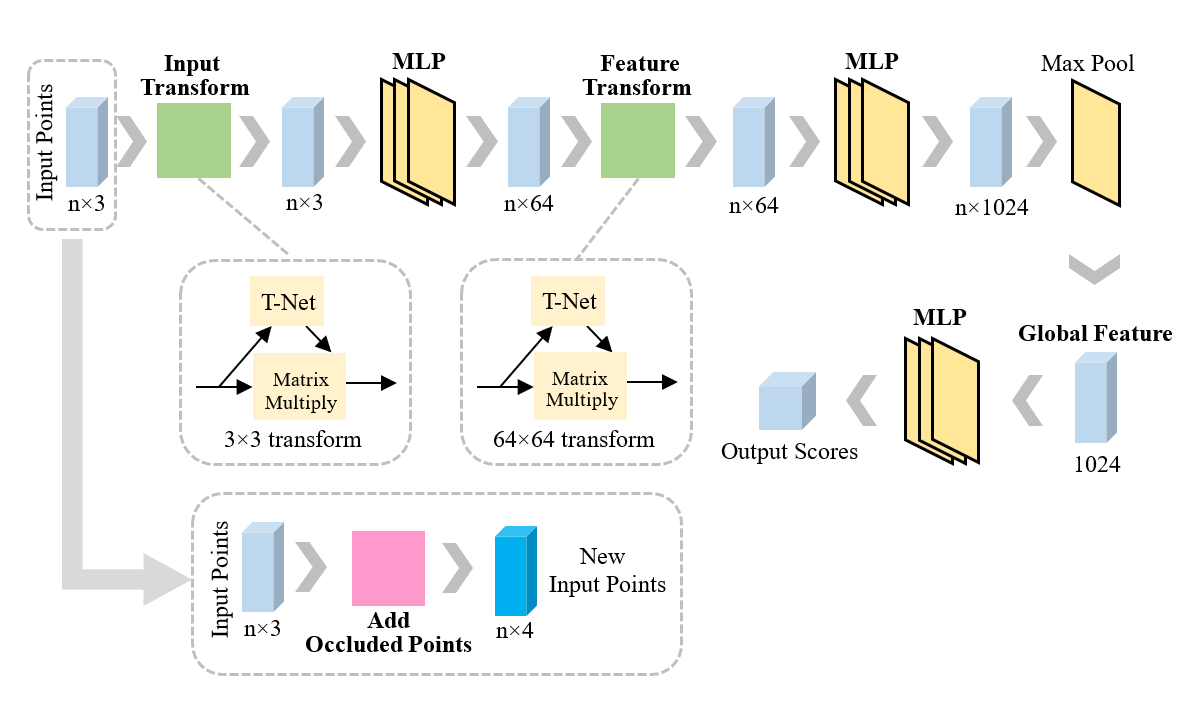}
\end{center}
   \caption{Here we show the main structure of the PointNet's classification network and the difference between the origin PointNet and ours. It is seen that we do not need to make many changes on the structure of the network itself.}
\label{fig:long}
\label{fig:onecol}
\end{figure*}

In Fig.5, we show the comparison of PointNet and our modified PointNet. The top figure is the original PointNet. The bottom figure is our modified PointNet. Changed parts are shown in the bottom bounding box. We can see that we do not need to make many changes on the structure of the network itself. Our approach could be applied to any network which can directly process the raw lidar point cloud data.

\section{Experimental Results}
We divide our experiments into two parts and we choose to perform experiments on the KITTI dataset. We firstly did experiments on the seven categories (`car', `van', `truck', `pedestrain', `cyclist', `tram' and `misc') of KITTI dataset. As `car', `van' and `truck' share a lot of similarities, and in fact they all belong to the `vehicle' category, we then merge car, van and truck to a single category, and perform experiments on these five categories.
\subsection{Classification Results on the 7 Categories}
We separately train the PointNet network on the original point cloud and the point cloud with occluded points. The classification results are shown in Table.1 and Fig.6. Experimental results show that both the overall accuracy and per-class accuracy of our approach have a significant improvement compared with the original PointNet. 

\begin{table}[h]
\begin{center} 
\caption{Classification results on the KITTI 7 categories dataset.}
\newcommand{\tabincell}[2]{\begin{tabular}{@{}#1@{}}#2\end{tabular}}
\begin{tabular}{p{2cm}<{\centering}p{2cm}<{\centering}p{2.5cm}<{\centering}p{2.5cm}<{\centering}}
\toprule  
&\bf dataset &\tabincell{c}{\bf accuracy\\\bf avg. class} &\tabincell{c}{\bf accuracy\\\bf overall}\\
\midrule  
Ours &KITTI &\bf 0.784 &\bf 0.920\\
\bottomrule 
\end{tabular}
\end{center}
\end{table}
\begin{figure}
\begin{center}
\includegraphics[width=0.9\linewidth]{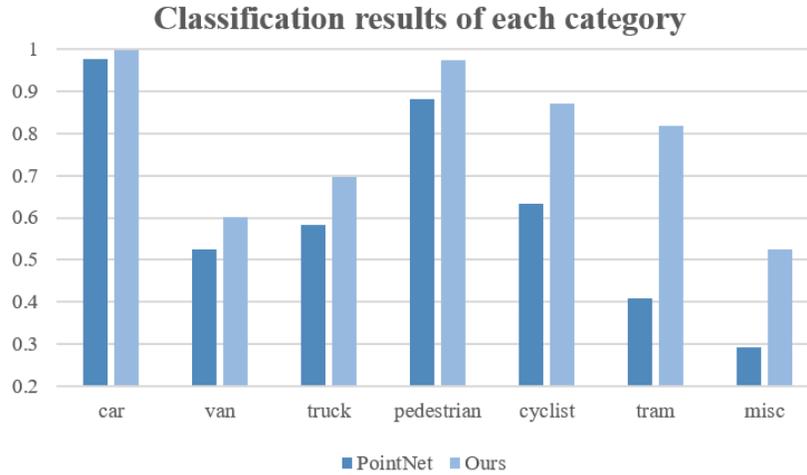}
\end{center}
   \caption{Classification results on the KITTI 7 categories dataset.}
\label{fig:long}
\label{fig:onecol}
\end{figure}

In Fig.7, we show the confusion matrix of the original PointNet and our approach. In Fig.8, we show the comparision between the point cloud with and without the added points. For many samples occluded by obstacles, their incomplete point cloud always result in wrong classification, such as sample C in Fig.8. Due to the incompleteness of the point cloud, sample C is classified as `misc' category in the original PointNet. In our approach, with the help of the added occluded points, it is correctly classified as the `car'.
\quad

\begin{figure}[H]
\centering
\subfigure[PointNet]{
\begin{minipage}[t]{0.5\linewidth}
\centering
\includegraphics[width=2.6in]{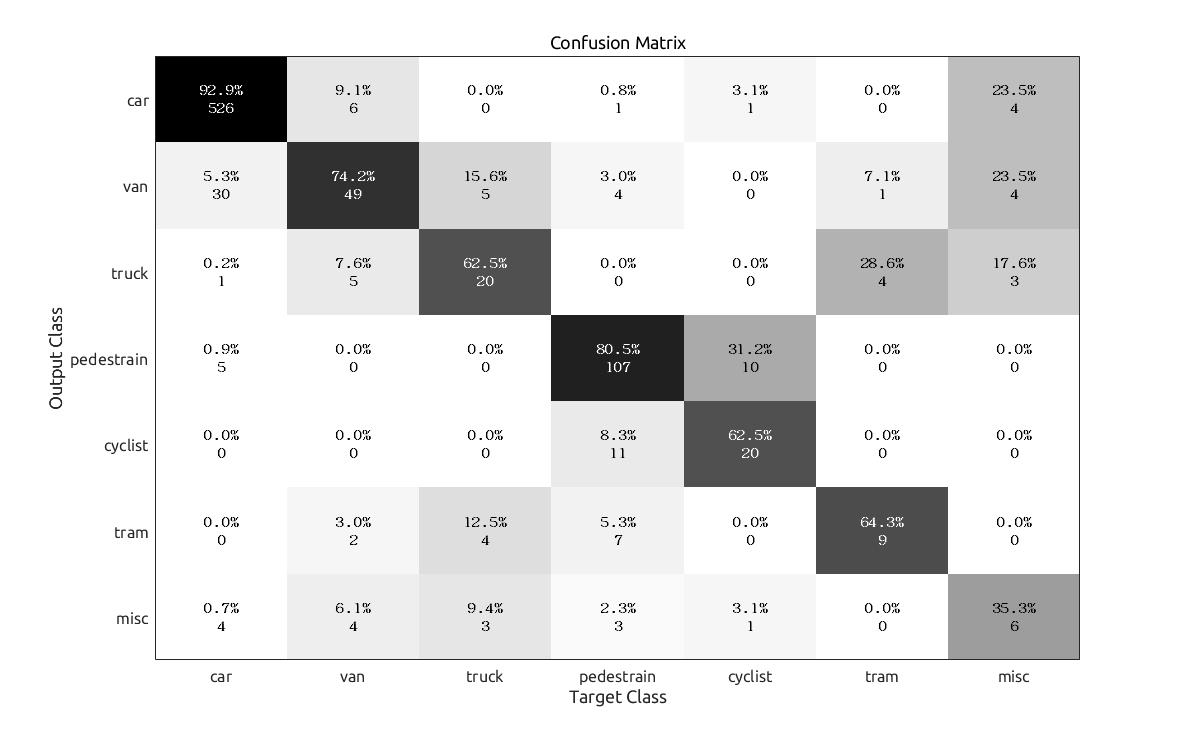}
\end{minipage}%
}%
\subfigure[Our Approach]{
\begin{minipage}[t]{0.5\linewidth}
\centering
\includegraphics[width=2.6in]{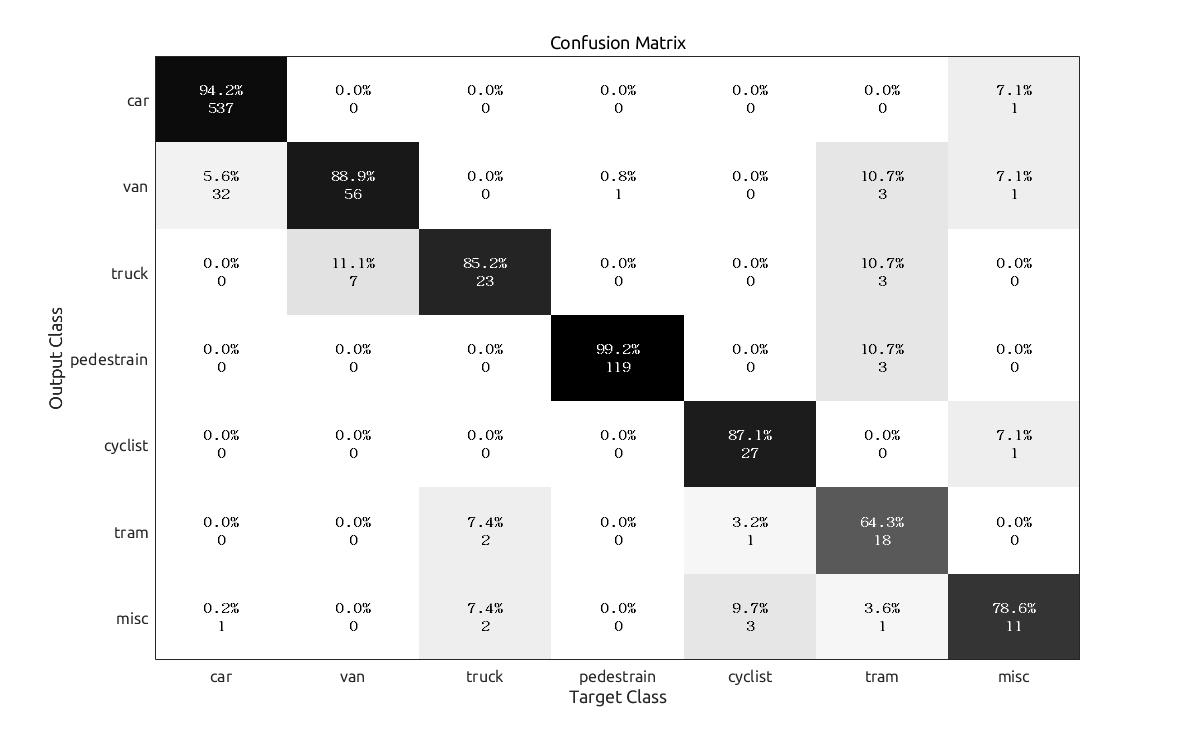}
\end{minipage}%
}%
\centering
\caption{Confusion matrix on the 7 categories using the original PointNet and our approach.}
\end{figure}

\begin{figure}[htbp]
\centering
\begin{minipage}[t]{0.15\linewidth}
\centerline{\includegraphics[width=0.6in]{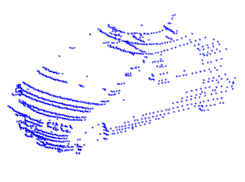}}
\end{minipage}
\quad
\begin{minipage}[t]{0.15\linewidth}
\centerline{\includegraphics[width=0.6in]{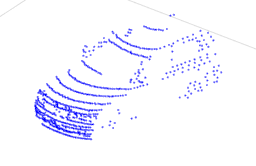}}
\end{minipage}
\quad
\begin{minipage}[t]{0.15\linewidth}
\centerline{\includegraphics[width=0.6in]{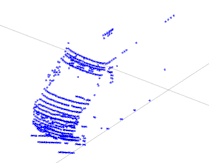}}
\end{minipage}
\quad
\begin{minipage}[t]{0.15\linewidth}
\centerline{\includegraphics[width=0.6in]{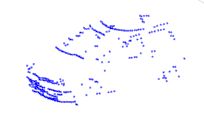}}
\end{minipage}
\quad
\begin{minipage}[t]{0.15\linewidth}
\centerline{\includegraphics[width=0.6in]{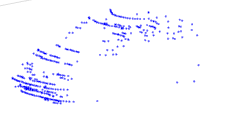}}
\end{minipage}
\quad

\begin{minipage}[t]{0.15\linewidth}
\centerline{\includegraphics[width=0.6in]{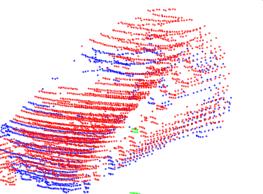}}
\centerline{(a)}
\end{minipage}
\quad
\begin{minipage}[t]{0.15\linewidth}
\centerline{\includegraphics[width=0.6in]{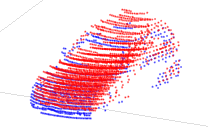}}
\centerline{(b)}
\end{minipage}
\quad
\begin{minipage}[t]{0.15\linewidth}
\centerline{\includegraphics[width=0.6in]{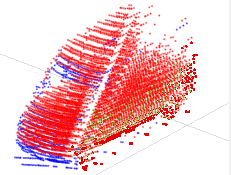}}
\centerline{(c)}
\end{minipage}
\quad
\begin{minipage}[t]{0.15\linewidth}
\centerline{\includegraphics[width=0.6in]{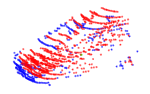}}
\centerline{(d)}
\end{minipage}
\quad
\begin{minipage}[t]{0.15\linewidth}
\centerline{\includegraphics[width=0.6in]{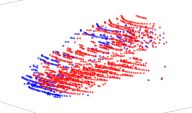}}
\centerline{(e)}
\end{minipage}
\centering
\caption{The original point cloud is colored in blue. The added occluded points are colored in red. The original point cloud is mostly occluded and may easily lead to a wrong classification ressult. With the help of the occluded points, these samples have now been correctly classified.}
\end{figure}

\subsection{Classification Results on the 5 Categories}
We merge car, van and truck into a single class and perform the experiments on the five categories. We believe that these three categories all belong to the `vehicle' class, and they are equally important to the self-driving cars. The classification results are shown in Table.3 and Fig.10.

\begin{table}[H]
\begin{center}
\caption{The percentage of each category's samples.} 
\newcommand{\tabincell}[2]{\begin{tabular}{@{}#1@{}}#2\end{tabular}}
\begin{tabular}{p{2.5cm}<{\centering}p{1.1cm}<{\centering}p{1.1cm}<{\centering}p{1.1cm}<{\centering}p{1.8cm}<{\centering}p{1.1cm}<{\centering}p{1.1cm}<{\centering}p{1.1cm}<{\centering}}
\toprule  
 &\bf car &\bf van &\bf truck &\bf pedestrain &\bf cyclist &\bf tram &\bf misc\\
\midrule  
\bf Testing data &0.626 &0.108 &0.038 &0.142 &0.036 &0.026 & 0.024\\
\bottomrule 
\end{tabular}
\end{center}
\end{table}

\begin{figure}[H]
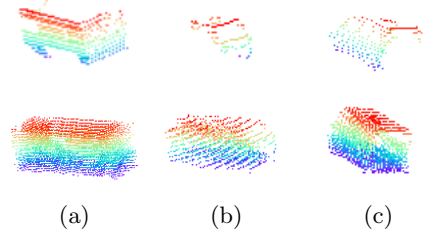

\centering
\begin{minipage}[t]{0.13\linewidth}
\centerline{\includegraphics[width=0.8in]{011.png}}
\end{minipage}
\quad
\begin{minipage}[t]{0.13\linewidth}
\centerline{\includegraphics[width=0.8in]{031.png}}
\end{minipage}
\quad
\begin{minipage}[t]{0.13\linewidth}
\centerline{\includegraphics[width=0.8in]{041.png}}
\end{minipage}
\quad

\begin{minipage}[t]{0.13\linewidth}
\centerline{\includegraphics[width=0.8in]{012.png}}
\centerline{(a)}
\end{minipage}
\quad
\begin{minipage}[t]{0.13\linewidth}
\centerline{\includegraphics[width=0.8in]{032.png}}
\centerline{(b)}
\end{minipage}
\quad
\begin{minipage}[t]{0.13\linewidth}
\centerline{\includegraphics[width=0.8in]{042.png}}
\centerline{(c)}
\end{minipage}
\quad
\caption{The object point cloud with and without occluded points of the van and car. Sample A is a van. Sample B and C are cars.}
\end{figure}
\setlength{\belowcaptionskip}{-1cm} 
\begin{table}[h]
\begin{center}
\caption{Classification results on the KITTI 5 categories dataset. We can see that the accuracy overall results of ours modified PointNet have better performance than PointNet.} 
\newcommand{\tabincell}[2]{\begin{tabular}{@{}#1@{}}#2\end{tabular}}
\begin{tabular}{p{2cm}<{\centering}p{2cm}<{\centering}p{2.5cm}<{\centering}p{2.5cm}<{\centering}}
\toprule  
&\bf dataset &\tabincell{c}{\bf accuracy\\\bf avg. class} &\tabincell{c}{\bf accuracy\\\bf overall}\\
\midrule  
Ours &KITTI &\bf 0.808 &\bf 0.962\\
\bottomrule 
\end{tabular}
\end{center}
\end{table}

In Fig.10, it is easily seen that each category's classification accuracy of our approach is improved in our approach. Some qualitative examples are shown in Fig.9. The confusion matrix is shown in Fig.11.

\begin{figure}[H]
\begin{center}
\includegraphics[width=0.9\linewidth]{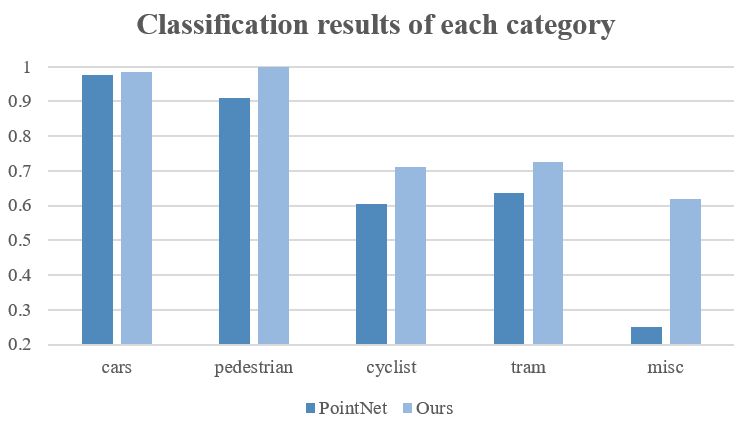}
\end{center}
\caption{Classification results on the KITTI 5 categories dataset.}
\label{fig:long}
\label{fig:onecol}
\end{figure}

\begin{figure}[H]
\centering
\subfigure[PointNet]{
\begin{minipage}[t]{0.5\linewidth}
\centering
\includegraphics[width=2.6in]{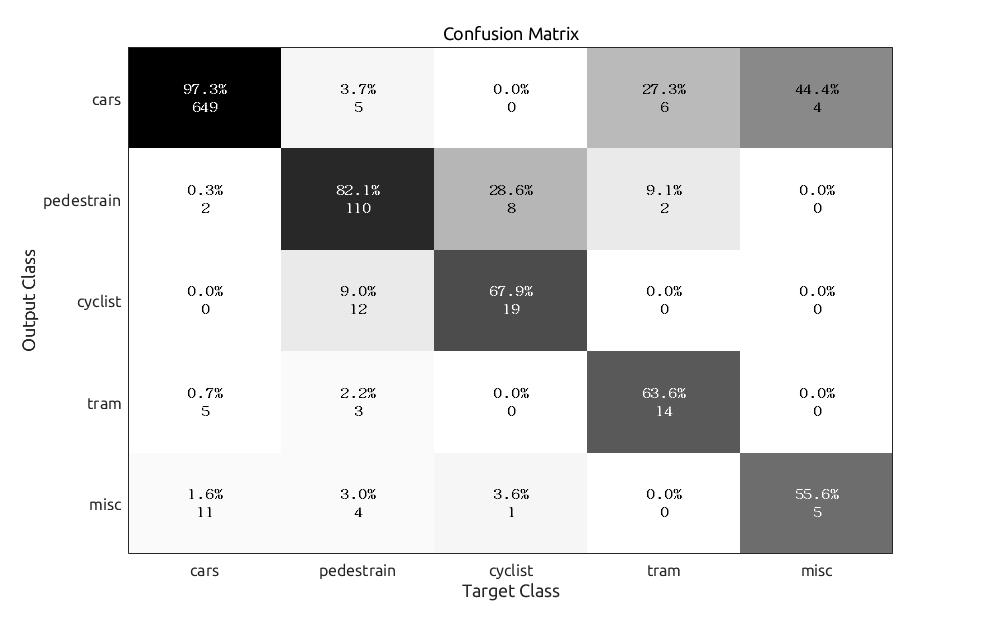}
\end{minipage}%
}%
\subfigure[Our Approach]{
\begin{minipage}[t]{0.5\linewidth}
\centering
\includegraphics[width=2.6in]{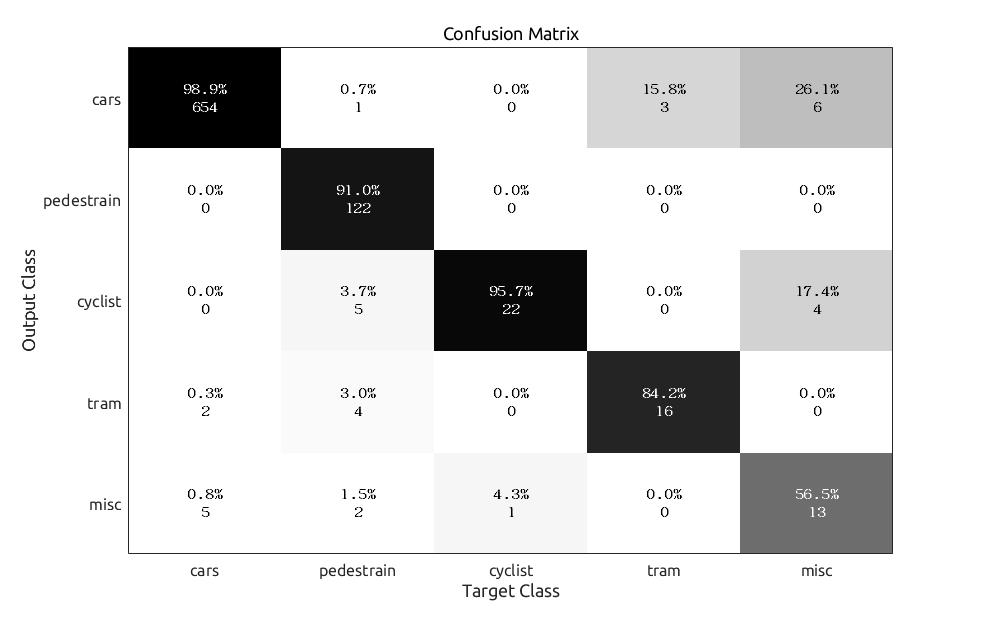}
\end{minipage}%
}%
\centering
\caption{Confusion matrix on the KITTI 5 categories using the original PointNet and our approach.}
\end{figure}
\quad

\section{Concluding Remarks}
In this paper, we investigate the lidar classification problem in occluded scenarios. We model occlusion as a intrinsic property of the lidar point cloud, and add a pre-precessing step to the lidar point cloud processing pipeline. It is important to emphasize that our approach is not only applicable to enhance PointNet's classification performance. We believe that our approach for modeling occlusion is an important pre-processing step that can enhance any classification approaches.

\bibliographystyle{splncs04}
\bibliography{reference}

\begin{thebibliography}{10}

\bibitem{roth1982ray}
Scott~D Roth.
\newblock Ray casting for modeling solids.
\newblock {\em Computer graphics and image processing}, 18(2):109--144, 1982.

\bibitem{qi2017pointnet}
Charles~R Qi, Hao Su, Kaichun Mo, and Leonidas~J Guibas.
\newblock Pointnet: Deep learning on point sets for 3d classification and
  segmentation.
\newblock In {\em Proceedings of the IEEE Conference on Computer Vision and
  Pattern Recognition}, pages 652--660, 2017.

\bibitem{petrovskaya2009model1}
Anna Petrovskaya and Sebastian Thrun.
\newblock Model based vehicle detection and tracking for autonomous urban
  driving.
\newblock {\em Autonomous Robots}, 26(2-3):123--139, 2009.

\bibitem{petrovskaya2009model}
Anna Petrovskaya and Sebastian Thrun.
\newblock Model based vehicle tracking in urban environments.
\newblock In {\em IEEE International Conference on Robotics and Automation,
  Workshop on Safe Navigation}, volume~1, pages 1--8, 2009.

\bibitem{petrovskaya2009efficient}
Anna Petrovskaya and Sebastian Thrun.
\newblock Efficient techniques for dynamic vehicle detection.
\newblock In {\em Experimental Robotics}, pages 79--91. Springer, 2009.

\bibitem{himmelsbach2009real}
Michael Himmelsbach, Thorsten Luettel, and H-J Wuensche.
\newblock Real-time object classification in 3d point clouds using point
  feature histograms.
\newblock In {\em 2009 IEEE/RSJ International Conference on Intelligent Robots
  and Systems}, pages 994--1000. IEEE, 2009.

\bibitem{wang2003online}
Chieh-Chih Wang, Charles Thorpe, and Sebastian Thrun.
\newblock Online simultaneous localization and mapping with detection and
  tracking of moving objects: Theory and results from a ground vehicle in
  crowded urban areas.
\newblock In {\em 2003 IEEE International Conference on Robotics and Automation
  (Cat. No. 03CH37422)}, volume~1, pages 842--849. IEEE, 2003.

\bibitem{wojke2012moving}
Nicolai Wojke and Marcel H{\"a}selich.
\newblock Moving vehicle detection and tracking in unstructured environments.
\newblock In {\em 2012 IEEE International Conference on Robotics and
  Automation}, pages 3082--3087. IEEE, 2012.

\bibitem{cheng2014robust}
Jian Cheng, Zhiyu Xiang, Teng Cao, and Jilin Liu.
\newblock Robust vehicle detection using 3d lidar under complex urban
  environment.
\newblock In {\em 2014 IEEE International Conference on Robotics and Automation
  (ICRA)}, pages 691--696. IEEE, 2014.

\bibitem{chen2017multi}
Xiaozhi Chen, Huimin Ma, Ji~Wan, Bo~Li, and Tian Xia.
\newblock Multi-view 3d object detection network for autonomous driving.
\newblock In {\em Proceedings of the IEEE Conference on Computer Vision and
  Pattern Recognition}, pages 1907--1915, 2017.

\bibitem{zhou2018voxelnet}
Yin Zhou and Oncel Tuzel.
\newblock Voxelnet: End-to-end learning for point cloud based 3d object
  detection.
\newblock In {\em Proceedings of the IEEE Conference on Computer Vision and
  Pattern Recognition}, pages 4490--4499, 2018.

\bibitem{Wang_2018_CVPR}
Xinlong Wang, Tete Xiao, Yuning Jiang, Shuai Shao, Jian Sun, and Chunhua Shen.
\newblock Repulsion loss: Detecting pedestrians in a crowd.
\newblock In {\em The IEEE Conference on Computer Vision and Pattern
  Recognition (CVPR)}, June 2018.

\bibitem{Zhang_2018_ECCV}
Shifeng Zhang, Longyin Wen, Xiao Bian, Zhen Lei, and Stan~Z. Li.
\newblock Occlusion-aware r-cnn: Detecting pedestrians in a crowd.
\newblock In {\em The European Conference on Computer Vision (ECCV)}, September
  2018.

\bibitem{Baque_2017_ICCV}
Pierre Baque, Francois Fleuret, and Pascal Fua.
\newblock Deep occlusion reasoning for multi-camera multi-target detection.
\newblock In {\em The IEEE International Conference on Computer Vision (ICCV)},
  Oct 2017.

\bibitem{Edward_2014}
Hsiao Edward and Hebert Martial.
\newblock Occlusion reasoning for object detectionunder arbitrary viewpoint.
\newblock In {\em IEEE Transactions on Pattern Analysis and Machine
  Intelligence}, pages 1803 -- 1815, 2014.

\bibitem{CTT}
{\em 3D LIDAR-based Dynamic Vehicle Detection and Tracking}.
\newblock PhD thesis, 2016.

\end{thebibliography}

\end{document}